\documentclass{article}

\usepackage{arxiv}

\usepackage[utf8]{inputenc} 
\usepackage[T1]{fontenc}    
\usepackage{hyperref}       
\usepackage{url}            
\usepackage{booktabs}       
\usepackage{amsfonts}       
\usepackage{nicefrac}       
\usepackage{microtype}      
\usepackage{lipsum}
\usepackage{graphicx}
\usepackage{lineno,hyperref}
\usepackage{bm}
\usepackage{caption}
\usepackage{multirow}
\usepackage[sort&compress,numbers]{natbib}
\emergencystretch=\maxdimen
\title{Robust Self-Supervised Convolutional Neural Network for Subspace Clustering and Classification}

\author{Ivica Kopriv$\mathrm{a}^*$}

\author{
 Dario Sitnik \\
  Laboratory for Machine Learning and Knowledge Representation \\ Division of Electronics \\ Ruđer Bošković Institute\\ Bijenička cesta 54, 10000, Zagreb, Croatia\\
  \texttt{dsitnik@irb.hr, dario.sitnik@gmail.com}
   \And
 Ivica Kopriva \\
  Laboratory for Machine Learning and Knowledge Representation \\ Division of Electronics \\ Ruđer Bošković Institute\\ Bijenička cesta 54, 10000, Zagreb, Croatia\\
  \texttt{ikopriva@irb.hr, ikopriva@gmail.com}
  }

\begin{document}
\maketitle
\begin{abstract}
Insufficient capability of existing subspace clustering methods to handle data coming from nonlinear manifolds, data corruptions, and out-of-sample data hinders their applicability to address real-world clustering and classification problems. This paper proposes the robust formulation of the self-supervised convolutional subspace clustering network ($S^2$ConvSCN) that incorporates the fully connected (FC) layer and, thus, it is capable for handling out-of-sample data by classifying them using a softmax classifier. $S^2$ConvSCN clusters data coming from nonlinear manifolds by learning the linear self-representation model in the feature space. Robustness to data corruptions is achieved by using the correntropy induced metric (CIM) of the error. Furthermore, the block-diagonal (BD) structure of the representation matrix is enforced explicitly through BD regularization. In a truly unsupervised training environment, Robust $S^2$ConvSCN outperforms its baseline version by a significant amount for both seen and unseen data on four well-known datasets. Arguably, such an ablation study has not been reported before.
\end{abstract}


\section{Introduction}
\label{sec:intro}
Up to now, it is recognized that many high-dimensional data can be modeled as samples drawn from a union of multiple low-dimensional subspaces. Examples that can be represented by subspaces include motion trajectories in a video \cite{costeira1998multibody}, face images \cite{basri2003lambertian} and hand-written digits \cite{hastie1997metrics}, to name a few. That is, each subspace corresponds to a class or category. The problem of grouping data according to the linear low-dimensional subspaces they are drawn from is known as subspace clustering \cite{vidal2011subspace}. Subspace clustering is achieved in two steps: (i) learning representation matrix \textbf{C} from data \textbf{X} and building corresponding affinity matrix $\mathbf{A}=|\mathbf{C}| + |\mathbf{C}^T|$; (ii) clustering the data into $k$ clusters by grouping the eigenvectors of the graph Laplacian matrix \textbf{L} that correspond with the leading $k$ eigenvalues. This second step is known as spectral clustering \cite{ng2002spectral,von2007tutorial}. Owning to the presumed subspace structure, the data points obey the self-expressiveness or self-representation property \cite{elhamifar2013sparse,peng2016constructing,liu2012robust,li2016structured,favaro2011closed}. In other words, each data point can be represented as a linear combination of other points in a dataset: \textbf{X=XC}. 

Subspace clustering approaches have achieved encouraging performance when compared with the clustering algorithms that rely on proximity measures between data points. However, they are still facing serious limitations when it comes to application to real-world problems. One limitation relates to the linearity assumption of the self-representation. That is because samples lie in non-linear subspaces in a wide range of applications, e.g. face images acquired under non-uniform illumination and different poses \cite{ji2017deep}. Standard practice for handling data from nonlinear manifolds is to use the kernel trick on samples mapped implicitly into high dimensional space. Therein, samples better conform to linear subspaces \cite{patel2013latent,patel2014kernel,xiao2015robust,brbic2018multi}. However, identifying an appropriate kernel function for a given data set is quite a difficult task \cite{zhang2019neural}. 

Motivated by the exceptional ability of deep neural networks to capture complex underlying structures of data and learn discriminative features for clustering \cite{hinton2006reducing,dilokthanakul2016deep,ghasedi2017deep,tian2014learning,xie2016unsupervised}, deep subspace clustering approaches emerged recently \cite{ji2017deep,abavisani2018deep,peng2016deep,yang2019deep,zhou2018deep,ji2019invariant,peng2018structured,peng2017cascade,zhou2019latent,zhang2019self,kheirandishfard2020multi}. In particular, it is shown that convolutional neural networks (CNNs), when applied to images of different classes, can learn features that lie in a union of linear subspaces \cite{lezama2018ole}. However, a large number of labeled data is required to train CNNs and that is often unavailable in practical applications. That, as an example, stands for one of the major problems in computational pathology \cite{colling2019artificial,abels2019computational} and field of medical imaging in general. In this regard, recent efforts in the development of deep subspace clustering methods are focused on networks that jointly learn feature representations, self-expressive coefficients and clustering results to produce more accurate clusters \cite{zhang2019self,kheirandishfard2020multi}. These approaches used pseudo labels generated by the spectral clustering module on currently learned representation to refine clustering accuracy in the next iteration. In label-constrained applications, that enables deep networks to learn representation from available unlabeled data and to use the small number of labeled data to validate the trained model. In this regard, the Self-Supervised Convolutional Subspace Clustering Network ($S^2$ConvSCN) \cite{zhang2019self} is of particular importance as it simultaneously trains the deep subspace clustering network and the fully connected (FC) layer with softmax classifier. That, in principle, solves the out-of-sample problem which is one of the limitations that prevent most of the existing subspace clustering methods from being applied to real-world problems. One notable exception in this regard is given in \cite{de2016kernel}. In $S^2$ConvSCN, a trained encoder and fully connected layer with a softmax classifier can be applied to test (unseen) data. 

However, challenges for the $S^2$ConvSCN are regularization for representation matrix \textbf{C}, robustness to errors in data, and the amount of prior information required to train the network. All of them individually and jointly limit the accuracy of the $S^2$ConvSCN type of network and, consequently, the capability to address real-world problems such as, for example, pixel-wise diagnosis of cancer from images of the histopathological specimens \cite{sitnik2020transfer}. Regularization of representation matrix \textbf{C} is imposed to ensure structural properties expected or known to hold in the subspace clustering problem. Among many regularizations low-rankness, \cite{liu2012robust,favaro2011closed} and sparseness \cite{elhamifar2013sparse}, of \textbf{C} are most successful in many applications. Low-rankness captures the global property of data while sparseness captures local property of data. These constraints guarantee exact clustering when subspaces are independent. However, independency assumption is overly restrictive for many real-world data sets \cite{tang2014structure,tang2016subspace}. This restriction is alleviated by combining low-rankness and sparsity \cite{li2016structured,brbic2018l0}. However, selection of Schatten-p ($S_p$) norm for measuring low-rankness and $\ell_p$-norm for measuring sparseness, where $0 \leq p \leq 1$, is a non-trivial problem. Use of $S_1$ - and/or $\ell_1$-norms yields convex optimization sub-problems but sacrifices accuracy because $S_1$ - and $\ell_1$-norms are not exact measures of rank and sparsity. As shown in \cite{brbic2018l0}, $S_0$ - and $\ell_0$ quasi-norm based regularization substantially improves the accuracy of low-rank sparse subspace clustering. The selection of $p\in[0, 1]$ in  $S_p/\ell_p$ regularization is however demanding. Results in \cite{zheng2017does} show that in $\ell_p$ regularized least squares  smaller values of $p$ lead to more accurate solutions. However, for the large noise the $\ell_1$-norm outperforms the $\ell_p$, $p<1$, norm. The recent result, based on replica analysis \cite{bereyhi2019statistical}, confirms the superiority of the $\ell_0$ quasi-norm over the $\ell_1$ norm in the noisy scenario but only if the regularization constant is chosen properly. Thus, the choice of the norm is noise-level dependent. However, by construction, representation matrix \textbf{C} has block diagonal (BD) structure. That is, multiplicity of $k$ of the eigenvalue 0 of the corresponding Laplacian matrix \textbf{L} equals the number of blocks in \textbf{C}, see Theorem 4 in \cite{lu2018subspace} and Proposition 4 in \cite{von2007tutorial}. Thus, it is natural to consider the BD regularization on \textbf{C} defined as a sum of $k$ smallest eigenvalues of \textbf{L} \cite{lu2018subspace,yin2015laplacian,khan2019approximate}. That becomes even more meaningful knowing that only regularizations based on selected $S_p$ and/or $\ell_p$ norms satisfy the enforced BD condition and have a direct connection with the BD property of the representation matrix, see Theorem 2 in \cite{lu2018subspace}.  

As noticed in \cite{lee2015membership,lee2018nonparametric}, the presence of noise and outliers moves \textbf{C} away from a BD structure. How to handle the errors (noise and corruptions) that possibly exist in data is yet another challenge that limits the capability of many existing subspace clustering algorithms to address real-world problems mentioned previously - the  $S^2$ConvSCN included. Many algorithms for shallow and/or deep subspace clustering assume specific type of errors related to noise, random corruptions or sample-specific corruptions \cite{elhamifar2013sparse,peng2016constructing,liu2012robust,li2016structured,favaro2011closed,xiao2015robust,tang2014structure,tang2016subspace,yin2015laplacian,zhang2019deep}. Some subspace clustering algorithms even do not take into account the possibility of data corruption \cite{ji2017deep,patel2013latent,patel2014kernel,brbic2018multi,zhang2019neural,dilokthanakul2016deep,ghasedi2017deep, xie2016unsupervised,ji2017deep,peng2016deep,yang2019deep,zhou2018deep,ji2019invariant,peng2018structured,peng2017cascade,zhou2019latent,zhang2019self,kheirandishfard2020multi,lezama2018ole,brbic2018l0,lu2018subspace,zhang2019convolutional,feng2014robust}. In real-world applications, learning subspace representation and handling errors present in data has to be solved together. That is because the ground-truth subspace information is often hidden behind the errors \cite{he2015information}. The solution to this problem is of high importance for networks of the type of the $S^2$ConvSCN \cite{zhang2019self,kheirandishfard2020multi,zhang2019convolutional}. That is because they use pseudo-labels generated by the subspace clustering network to train the FC layer used for classification of the unseen data. The problem with error specific models is that errors in real data have different origins and magnitudes and therefore may not follow the presumed model. For many data, it is hard to know precisely what are the origins of data corruptions. It is also known that errors in real data very often have impulsive character and that makes the mean square error (MSE) loss function, commonly used in deep learning and subspace clustering, a suboptimal choice from the robustness point of view \cite{liu2007correntropy,chen2016efficient}. That is why correntropy induced loss function, also known as correntropy induced metric (CIM) \cite{liu2012robust}, is used for robust deep learning \cite{chen2016efficient,heravi2018new,qi2014robust} and/or subspace clustering \cite{he2015information,xing2019correntropy,zhang2019robust,jin2018correntropy}. Correntropy is derived from Renyi's quadratic entropy with the Parzen window estimation. Because it is estimated directly from data, correntropy can handle outliers under any distribution or the sample-specific errors in data. By mapping the input data space to a reproducible kernel Hilbert space (RKHS), correntropy defines an $\ell_2$ distance in RKHS and creates nonlinear distance measure in the original input data space \cite{gunduz2009correntropy}. Thus, CIM is actually MSE in RKHS. In a nutshell, MSE implies Gaussian or normal distributions of errors while correntropy handles non-Gaussian distribution of error in a data-driven manner.  

Motivated by discussed limitations of (deep) subspace clustering methods to address real-world problems such as robustness to errors and handling out-of-sample (unseen) data this paper enlists five main contributions:    
\begin{enumerate}
    \item to improve the robustness of the $S^2$ConvSCN network and its performance on unseen data, we propose the CIM instead of the MSE in the self-expression module in the feature space of the $S^2$ConvSCN. Minimization of the CIM in a data-driven manner maximizes the probability of error (discrepancy between data and its self-representation) to be zero for an arbitrary distribution. 
    \item we propose BD regularization in learning the representation matrix \textbf{C} in the self-expression module in feature space of the $S^2$ConvSCN. As opposed to \cite{lu2018subspace,yin2015laplacian,khan2019approximate,zhang2019convolutional,feng2014robust} that use indirect (implicit) measures of the block-diagonality of \textbf{C}, we define the BD regularization $\left \| \mathbf{C} \right \|_{[k]}$ as a sum of $k$ smallest eigenvalues of the Laplacian matrix \textbf{L}. By using the Lanczos algorithm, $k$ smallest eigenvalues are computed in a computationally efficient manner. As opposed to \cite{zhang2019convolutional}, where BD-regularization is used with the deep subspace clustering network \cite{ji2017deep} we use it with the $S^2$ConvSCN to improve its performance in label-free learning. 
    \item BD-regularization is integrated into end-to-end gradient-based learning in contrast to approach in \cite{zhang2019convolutional}.
    \item state-of-the-art shallow and deep subspace clustering methods are trained and tested on the whole dataset. That, up to a significant extent, limits their capability to handle the out-of-sample data. Also, violation of unsupervised learning principles by optimizing hyperparameters of the network using true labels excludes these algorithms from the fair comparison with the model presented in this study. Thus, we compare Robust $S^2$ConvSCN with $S^2$ConvSCN under the same conditions, i.e. on whole datasets without using labels during the model learning phase. 
    \item to validate the capability of Robust $S^2$ConvSCN and $S^2$ConvSCN to classify unseen data, we train and test models using independent folds (partitions) of the dataset. Depending on the size of each of the four datasets we use 5 or 15 folds. Thus, we conduct an ablation study where the individual and joint contribution of BD-regularization and CIM were compared with the baseline version of the $S^2$ConvSCN that uses MSE and $\ell_2$ norm of \textbf{C} in the self-expression module in the feature space. As can be seen from the results of the ablation study in section 5,  CIM based Robust $S^2$ConvSCN improves the performance over the baseline $S^2$ConvSCN on all datasets by a significant amount. Additionally, BD-regularization jointly with CIM leads to better performance than $\ell_2$ regularization on one dataset. Therefore, there is still a large space for performance improvement of the $S^2$ConvSCN type of the network on the classification of unseen data. To the best of our knowledge, such type of ablation study has not been reported previously.
    
\end{enumerate}

The rest of the paper is organized as follows. Section \ref{sec:backg} gives a brief overview of subspace clustering algorithms, outlines their limitations and proposes alternative formulations. Preliminaries related to BD regularization and CIM are given in Section \ref{sec:prelim}. Section \ref{sec:robusts2convscn} introduces improvements of the $S^2$ConvSCN in term of BD regularization and CIM based error measure. Comparative performance analysis of proposed Robust $S^2$ConvSCN is conducted in section \ref{sec:experimental}. Discussion and conclusion are presented in section \ref{sec:discussion}. 

\section{Background and related work}
\label{sec:backg}

\subsection{Main notations and definitions}

Throughout this paper, matrices are represented with bold capital symbols and vectors with bold lower-case symbols. $\mathbf{X} \in \mathbb{R}^{d \times N}$ represents data matrix comprised of $N$ data samples with dimensionality $d$. $\left \{ \mathbf{H}^{(l)}_i \right \}^{m^{(l)}}_{i=1}$ represent feature maps produced at the output of layer $l-1$. Thus, $\mathbf{H}^{(0)}=\mathbf{X}$ and $\mathbf{H}^{(L)}=\mathbf{\hat{X}}$. $\mathbf{\hat{X}}$ represents the output of the decoder and $L$ represents number of convolutional layers in the autoencoder. $\left \{ \mathbf{w}^{(l)}_i \right \}^{m^{(l)}}_{i=1}$ stand for a set of filters with associated biases $\left \{ \mathbf{b}^{(l)}_i \right \}^{m^{(l)}}_{i=1}$ that form a convolutional layer $l=1,\dots,L$. $\mathbf{z}_n = \left [ \mathbf{h}^{(L/2)}_1 \left ( : \right ) \dots \mathbf{h}^{(L/2)}_{m^{(L/2)}} \left ( : \right ) \right ]^T \in \mathbb{R}^{\hat{d} \times 1}$ stands for feature vector comprised of vectorized and concatenated feature maps, with $\hat{d}$ extracted features, in the top layer $\frac{L}{2}$ (encoder output) representing input sample $\mathbf{x}_n, n=1, \dots ,N$. $ \mathbf{Z} = \left [ \mathbf{z}_1 \dots \mathbf{z}_N \right ]$ represents feature matrix. $\mathbf{C} \in \mathbb{R}^{N \times N}$ stands for representation matrix in self-expressive model $\mathbf{Z=ZC}$. $\mathbf{A} = |\mathbf{C}| + |\mathbf{C}^T|$  is the affinity matrix and $\mathbf{L = D^{- \frac{1}{2}} A D^{\frac{1}{2}}}$ is corresponding graph Laplacian matrix. \textbf{D} is diagonal degree matrix such that $\mathbf{D}_{ii}=\sum^N_{j=1}\mathbf{A}_{ij}$. $\left \| \mathbf{X} \right \|_F= \sqrt{\sum^N_{i,j=1} x^2_{ij}} $ is the Frobenius norm of matrix \textbf{X}. $\ell_p(\mathbf{x})=\left \| \mathbf{x} \right \|^p_p = \sum^d_{i=1}\left \| x_i \right \|^p,\ 0 <p\leq 1$ is the $\ell_p$ norm of \textbf{x}. $\ell_0(\mathbf{x})=\left \| \mathbf{x} \right\|_0 = \#\{x_i \neq 0,\ i=1,\dots,d\}$, where \# denotes the cardinality function, is $\ell_0$ quasi norm of \textbf{x}. The $S_p$, $0 < p\leq 1$, Schatten norms of matrix \textbf{X} are defined as the corresponding $\ell_p$ norms of the vector of singular values of \textbf{X}, i.e. $S_p( \mathbf{X}) =\left \| \sigma (\mathbf{X}) \right\|_p^p$ where $\sigma(\mathbf{X})$ stands for the vector of singular values of \textbf{X}. Depending on the context, \textbf{0} represents matrix/vector of all zeros and \textbf{1} represents matrix/vector of all ones. Grouping the data according to the linear subspaces they are drawn from is known as subspace clustering \cite{vidal2011subspace}. The problem is formally defined in Definition 1.\bigbreak

\noindent \textbf{Definition 1.} Let $\mathbf{X}=[\mathbf{X}_1,\dots,\mathbf{X}_k]=[\mathbf{x}_1,\dots,\mathbf{x}_N]\in \mathbb{R}^{d \times N}$ be a set of sample vectors drawn from a union of $k$ subspaces in $\mathbb{R}^d ,\ \cup^k_{i=1} \left \{ S_i \right \}$, of dimensions $d_i \ll \min \left \{ d,N \right \}$ , for $i=1,\dots,k$. Let $\mathbf{X}_i$ be a collection of $N_i$ samples drawn from subspace $S_i,\  N=\sum^k_{i=1}N_i$. The problem of subspace clustering is to segment samples into the subspaces they are drawn from.\bigbreak
\noindent Throughout this paper, as it is the case in majority of other papers, we have assumed that number of clusters $k$ is known \textit{a priori}.

\subsection{Sparse and low-rank subspace clustering}
Many approaches to deep subspace clustering are based on the introduction of the self-representation in the feature space \cite{abavisani2018deep,ji2017deep,peng2016deep,zhou2018deep,zhou2019latent,zhang2019self,kheirandishfard2020multi,zhang2019convolutional}. Thus, by the virtue of this, improvements of the shallow subspace clustering methods are of direct relevance to their deep counterparts. Subspace clustering task is accomplished through (i) learning the representation matrix \textbf{C} from data \textbf{X}, and (ii) clustering the data into $k$ clusters by grouping the eigenvectors of the graph Laplacian matrix \textbf{L} that correspond with the $k$ leading eigenvalues. This second step is known as spectral clustering \cite{ng2002spectral,von2007tutorial}. Low-rank sparse subspace clustering aims to learn the low-rank sparse representation matrix by solving the following optimization problem\cite{li2016structured}: 

\begin{equation}
\label{Eq1}
  \min_{\mathbf{C}}\lambda \left \| \mathbf{C} \right \|_{S_p}+\tau \left \| \mathbf{C} \right \|^p_p\ s.t.\  \mathbf{Z=ZC+E},\ diag(\mathbf{C})=\mathbf{0}
\end{equation}

\noindent where $\lambda$ and $\tau$ are nonnegative regularization constants. If number of layers $L=0$ problem (\ref{Eq1}) is related to shallow subspace clustering. Constraint $diag(\mathbf{C})=\mathbf{0}$ is necessary to prevent sparseness regularized optimization algorithms to converge towards trivial solution where each data point represents itself. This constraint is not necessary for problem constrained only by low-rank. Low rank \cite{liu2012robust,li2016structured} and sparse \cite{elhamifar2013sparse} subspace clustering is obtained in (\ref{Eq1}) by setting $\tau=0$ and $\lambda=0$, respectively. When data samples are contaminated with additive white Gaussian noise (AWGN) problem (\ref{Eq1}) becomes: 

\begin{equation}
\label{Eq2}
  \min_{\mathbf{C}} \left \| \mathbf{E} \right \|^2_F+\lambda \left \| \mathbf{C} \right \|_{S_p}+\tau \left \| \mathbf{C} \right \|^p_p \ s.t.\ diag(\mathbf{C})=\mathbf{0}
\end{equation}

\noindent Alternatively, square of the Frobenius norm of \textbf{C} is used for regularization \cite{lu2012robust}:

\begin{equation}
\label{Eq3}
  \min_{\mathbf{C}} \left \| \mathbf{E} \right \|^2_F+\lambda \left \| \mathbf{C} \right \|^2_{F}
\end{equation}

\noindent Objective (\ref{Eq3}) is used also in the self-expression module of the $S^2$ConvSCN in \cite{zhang2019self}. As seen from (\ref{Eq2}) and (\ref{Eq3}), the MSE measure for discrepancy between \textbf{Z} and its self-representation \textbf{ZC} is justified only for the contamination by the AWGN. For sample-specific corruptions (outliers) the proper norm is $\left \| \mathbf{E} \right \|_{2,1}$ while for large random corruptions the proper choice is $\left \| \mathbf{E} \right \|_{1}$ \cite{liu2012robust}. However, errors in real world data have different origins and magnitude and may not follow specific probabilistic model. Sometimes, it is hard to know the true origin of corruptions present in data. Thus, to obtain method robust to arbitrary corruption we propose to introduce the CIM of the error. As discussed in the introduction, selection of $S_p$- and $\ell_p$ norm is nontrivial issue depending also on the amount of noise present in data. Rationale behind introduction of any regularization on \textbf{C} is to reflect its structural property of block-diagonality. Even though $\left \| \mathbf{C}\right \|_{S_p}$ and $\left \|\mathbf{C} \right \|_{p}$, $0 \leq p \leq 1$ in principle satisfy the enforced block-diagonality condition, their approximation of the BD structure of \textbf{C} is indirect \cite{lu2018subspace}. Hence, we propose to introduce BD regularization on representation matrix \textbf{C}.

\section{Preliminaries}
\label{sec:prelim}

\subsection{Block diagonal regularization}
To introduce BD regularization we state the proposition 4 from \cite{von2007tutorial} for the graph Laplacian
matrix \textbf{L}: \bigbreak

\noindent \textbf{Proposition 1.} (Proposition 4 in \cite{von2007tutorial}: Number of connected components and spectra of \textbf{L}). Let \textbf{G} be an undirected graph with nonnegative weights. Then the multiplicity $k$ of the eigenvalues 0 of \textbf{L} equals the number of connected components in the graph. \bigbreak

\noindent Thus, based on Proposition 1 we define the BD regularization of \textbf{C} as the sum of the $k$ smallest eigenvalues of \textbf{L}:

\begin{equation}
    \left \| \mathbf{C} \right \|_{[k]} = \sum^N_{i=N-k+1}\lambda_i(\mathbf{L})
\end{equation}
\bigbreak
\noindent where $\lambda_i(\mathbf{L})$, $\lambda_1 \geq \lambda_2 \geq \dots \geq \lambda_{(N-k+1)}\geq \dots \geq \lambda_N$, stands for the $i$-th eigenvalue of \textbf{L}.

\subsection{Correntropy}
Here we briefly introduce the correntropy and its properties that qualify it as loss function robust to data corruptions. Let $\mathbf{S}=[\mathbf{s}_1,\dots,\mathbf{s}_N]\in\mathbb{R}^{d \times N}$ and $\mathbf{T}=[\mathbf{t}_1,\dots,\mathbf{t}_N]\in\mathbb{R}^{d \times N}$ be two realizations of the corresponding random variables. The empirical correntropy is estimated from data as:

\begin{equation}
    \label{Eq5}
    \hat{V}(\mathbf{S},\mathbf{T})=\frac{1}{N}\sum^N_{i=1}\kappa_\sigma(\mathbf{s}_i,\mathbf{t}_i)
\end{equation}

\noindent where

\begin{equation}
    \kappa_\sigma(\mathbf{s}_i,\mathbf{t}_i)=\exp \left ( -\frac{\left \| \mathbf{s}_i - \mathbf{t}_i \right \|}{2\sigma^2} \right )
\end{equation}
\bigbreak
\noindent is the Gaussian kernel. Herein, we present from \cite{liu2007correntropy} two (out of ten) properties of correntropy  that justify its use as robust error measure. \bigbreak

\noindent \textbf{Property 1.} (Property 3 in \cite{liu2007correntropy}). Correntropy involves all the even moments of the random variable $\bm{\varepsilon}=\mathbf{T-S}$:

\begin{equation}
    V_\sigma (\mathbf{S},\mathbf{T}) = \frac{1}{\sqrt{2 \pi \sigma}} \sum^\infty_{n=0} \frac{(-1)^n}{2^n n!} \mathbb{E} \left [ \frac{(\mathbf{S}-\mathbf{T})^{2n}}{\sigma^{2n}} \right ]
\end{equation}
\bigbreak
\noindent where $\mathbb{E}$ denotes mathematical expectation. As $\sigma$ increases, the high-order moments decay faster; so, the second-order moment tends to dominate, and the correntropy approaches correlation. 

While the MSE involves only second-order moments and is, thus, optimal measure for error distributed normally, correntropy is the optimal measure for error with the arbitrary (non-Gaussian) distribution. Furthermore, as can be seen from Equation (\ref{Eq5}), correntropy is data-driven.\bigbreak

\noindent \textbf{Property 2} (Property 8 in \cite{liu2007correntropy}). The function:

\begin{equation}
    CIM (\mathbf{S},\mathbf{T}) = \left ( \kappa (\mathbf{0},\mathbf{0})-\hat{V}(\mathbf{S},\mathbf{T}) \right )^{\frac{1}{2}}
\end{equation}

\noindent defines a CIM in sample space.

\section{Robust self-supervised convolutional subspace clustering network}
\label{sec:robusts2convscn}

\begin{figure}[!ht]
\centering
\includegraphics[width=0.8\textwidth]{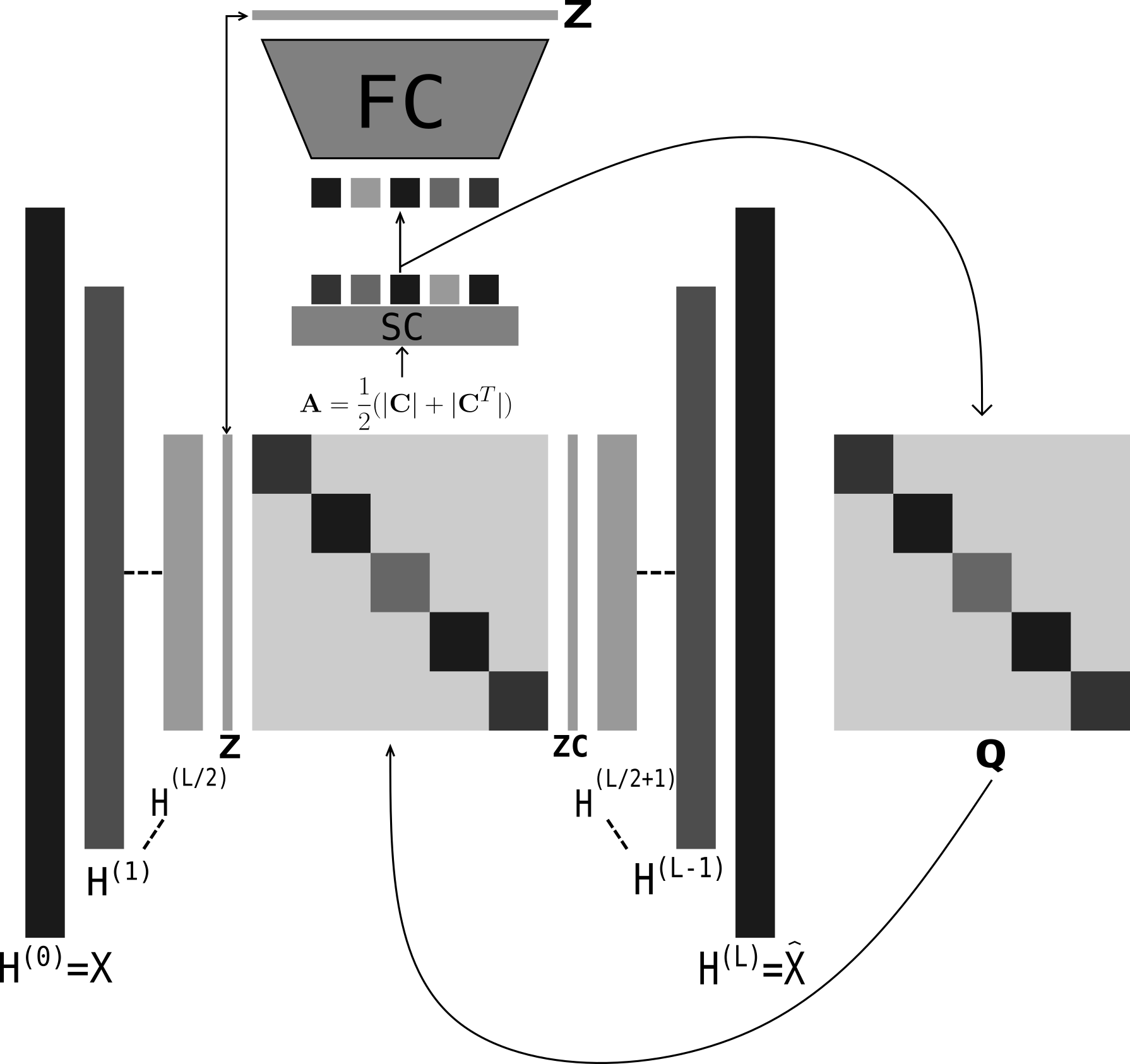}
\caption{Self-supervised $S^2$ConvSCN model showing all layers and modules. Encoder and decoder layers are denoted by $H^{(0)}$ to $H^{(L)}$. Together, flattened encoder features \textbf{Z} and representation matrix weights \textbf{C} construct self-expression layer \textbf{ZC} where optimal \textbf{C} is learned. Spectral clustering (SC) module receives affinity matrix \textbf{A} as an input. Pseudo-labels from SC module serve as target labels for the FC-softmax classifier. Also, pseudo-labels form a matrix \textbf{Q} where element $q_{ij}$ is set to 1 if samples $i$ and $j$ have the same pseudo-label. Otherwise, $q_{ij}$ is set to 0. The matrix \textbf{Q} is then used as a regularizer for the representation matrix \textbf{C} to learn a better representation.}
\label{figure:robust_s2convscn}
\end{figure}

We first provide a basic description of the $S^2$ConvSCN \cite{zhang2019self} and briefly discuss its weaknesses. The base of the $S^2$ConvSCN is autoencoder. It is a fully convolutional network (FCN) that tries to replicate the input to the output. The encoder performs dimensionality reduction to the latent code while the decoder expands that code to the output. The goal of the autoencoder is to get the output as close as it gets to the input data. An additional self-expression module, which is following the flattened latent code, is forming a DSCNet \cite{ji2017deep}. While DSCNet is producing representation matrix \textbf{C}, the clustering algorithm can only cluster the train data. Thus, it is not capable of dealing with the out-of-sample problem. Figure \ref{figure:robust_s2convscn} shows complete $S^2$ConvSCN architecture with all mentioned layers and modules.

One novelty that is presented in $S^2$ConvSCN is an FC layer, attached to the latent code, that is upgrading the DSCNet model. On its output, FC has softmax which can directly classify the input data. Upon finishing the training procedure, the encoder and FC layer form a new model that can be used for evaluating the learning procedure on the independent test set. While authors of the $S^2$ConvSCN model \cite{zhang2019self} provide a tool for dealing with the out-of-sample model, the actual performance on an independent test set has not been tested. 

The second novelty of $S^2$ConvSCN is training the FC layer from pseudo-labels. When training the network, the spectral clustering module clusters the data according to the affinity matrix every $E$ epochs and updates the pseudo-labels. It is important to note that the mentioned affinity matrix is constructed from the matrix \textbf{C} learned in the self-expression layer and it is changing every epoch during the training. Pseudo-labels assigned to the data are used in two ways. The first is to train the FC layer for classification, and the second is to suppress the \textbf{C} matrix values for samples that do not belong to the same cluster. Both ways are known as self-supervision.

Authors of both \cite{ji2017deep} and \cite{zhang2019self} suggest that, to converge, parts of the model should be pretrained before training the whole model. However, there is no guaranty that the model will keep converging after attaching final layers or modules. For example, if a pretrained DSCNet yields reasonably good pseudo-labels, it is possible that after attaching the FC layer and performing self-supervision in $S^2$ConvSCN, the matrix \textbf{C} will get worse and the whole model will diverge from the optimal solution. Thus, it is important to tune constants associated with the loss functions of the mentioned modules. Also, when training the neural networks, it is suggested to reduce the learning rate when the sum of all losses reaches a plateau. As the $S^2$ConvSCN has many loss functions regarding different layers or modules, smaller learning rate could be beneficial in the sense that the overall loss can reach a better minimum in the error space. While the smaller learning rate on the plateau is helpful to the most-contributing losses, it could slow down the convergence of other losses. Arguably, the same thing would happen in a model with many different loss functions such as $S^2$ConvSCN. Thus, $S^2$ConvSCN is not resistant to spoiling the goodness of representation matrix \textbf{C} while trying to reach the loss minimum. 

\begin{figure}[ht!]
    \centering
    \includegraphics[width=0.8\textwidth]{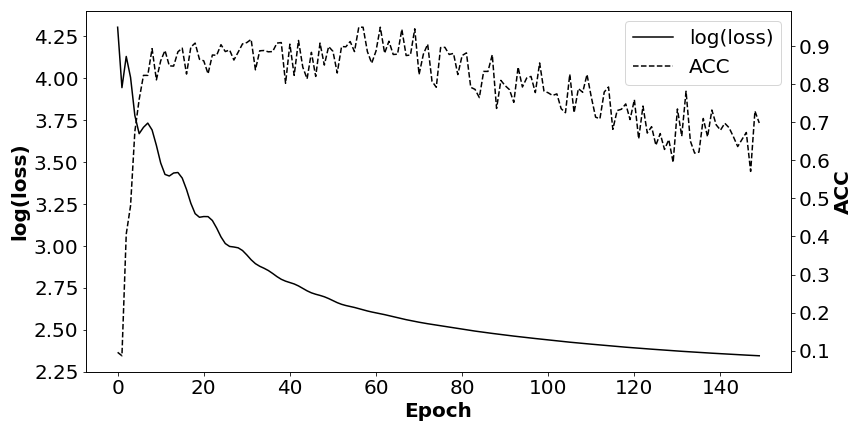}
    \includegraphics[width=0.8\textwidth]{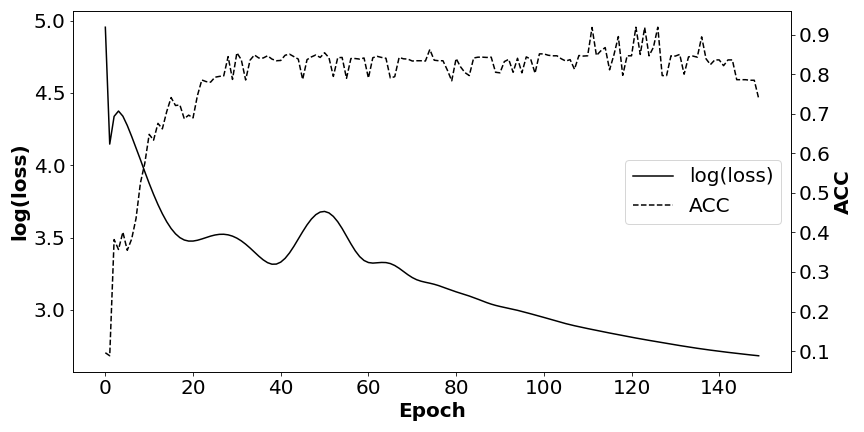}
    \caption{Figures illustrate the change of accuracy and loss over epochs for two DSCNet models. The top figure is showing the DSCNet model with MSE in a self-expressive model, while the bottom figure represents Robust DSCNet with CIM in the self-expressive model. Both models were trained and tested on a full COIL20 dataset with $\ell_2$ regularization of representation matrix \textbf{C} (see Equation (\ref{Eq10})). It can be seen that accuracy is oscillating even though loss is decreasing. Thus, selecting the model when the accuracy is highest makes an unfair practice.}
    \label{fig:acc_vs_loss}
\end{figure}

According to the training procedure given in \cite{jiGithub}, the performance of the DSCNet strongly depends on observing true labels every epoch. Although presented as an unsupervised method, DSCNet is using the minimum of clustering error as an early stopping criterion which is in contradiction with the principles of unsupervised learning and arguably leads to the overfitting of the model. As can be seen in Figure \ref{fig:acc_vs_loss}, much better representation matrix \textbf{C} can be learned from a model by ignoring loss and stopping when the accuracy is highest. In addition to representing supervised intervention in the learning process, it also yields overly optimistic performance. That is because there is no way to test the network on an independent test set. 

Another problematic part is that the DSCNet algorithm is post-processing matrix \textbf{C} with a hyper-parameter optimized using the ground truth labels. Keeping in mind that: the $S^2$ConvSCN is upgraded DSCNet model, it is using the pretrained DSCNet parts of the network, and it has not been tested on the independent dataset, it is reasonable to conclude that the label-leakage also appeared in $S^2$ConvSCN. Regardless of having the self-supervision modules, the first pseudo-labels generated from DSCNet contain leaked knowledge about the group affiliations of the data.

Based on the previous discussion, we propose two new objective functions $\mathcal{L}_{CIM}$ for the self-expression module of the $S^2$ConvSCN:

\begin{equation}
    \label{Eq9}
    \min_\mathbf{C} \mathrm{CIM}^2(\mathbf{E})+\gamma \left \| \mathbf{C} \right \|_{[k]} , 
\end{equation}
\begin{equation}
    \label{Eq10}
    \min_\mathbf{C} \mathrm{CIM}^2(\mathbf{E})+\gamma \left \| \mathbf{C} \right \|_2 
\end{equation}

\noindent where $\mathbf{E}=\mathbf{Z}-\mathbf{ZC}$. \textbf{Z} and \textbf{ZC} stand for input and output features of self-expression layer, respectively. $\left \| \mathbf{C} \right \|_{[k]}$ denotes BD regularization and $\left \| \mathbf{C} \right \|_2$ denotes $\ell_2$ regularization of representation matrix \textbf{C}. $\gamma$ represents a trade-off constant. The objectives (\ref{Eq9}) and (\ref{Eq10}) ensure a smooth decrease of the loss function that enables the use of label-free stopping criterion. For the sake of completeness, other loss functions of Robust $S^2$ConvSCN are stated. Auto-encoder reconstruction loss, where \textbf{X} represents the input data and $\mathbf{\hat{X}}$ represents the output of the decoder (also $\mathbf{H}^{(0)}$ and $\mathbf{H}^{(L)}$ in Figure \ref{figure:robust_s2convscn}, respectively), is defined as:

\begin{equation}
    \label{eq:rec_loss}
    \mathcal{L}_{REC} = \frac{1}{2N}\sum^N_{j=1} \left \| \mathbf{x}_j - \hat{\mathbf{x}}_j \right \|^2_2 =  \frac{1}{2N} \left \| \mathbf{X} - \mathbf{\hat{X}} \right \|^2_F
\end{equation}

\noindent When matrix \textbf{Q} is calculated, (see Figure \ref{figure:robust_s2convscn}), $\left \| \mathbf{C} \right \|_\mathbf{Q}$ loss is defined as:

\begin{equation}
    \label{eq:cq_loss}
    \mathcal{L}_{CQ} =\sum_{i,j} |c_{ij}|\frac{\left \| \mathbf{q}_i - \mathbf{q}_j \right \|^2_2}{2} := \left \| \mathbf{C} \right \|_\mathbf{Q}
\end{equation}

\noindent where $\mathbf{q}_i$ and $\mathbf{q}_j$ represent one-hot encoded pseudo-labels for $i$-th and $j$-th sample. $|c_{ij}|$ stands for the absolute representation value for the  corresponding samples in the representation matrix \textbf{C}. Cross-entropy loss of FC layer is:

\begin{equation}
    \label{eq:cross_ent}
    \mathcal{L}_{CE} =\frac{1}{N}\sum_{j=1}^N ln(1+e^{\hat{\mathbf{y}}_j^T \mathbf{q}_j})
\end{equation}

\noindent where $\hat{\mathbf{y}}_j^T$ stands for a softmax normalized output of the FC layer for $j$-th sample. Furthermore, a center loss is applied to the logits $\mathbf{y}_j$:

\begin{equation}
    \label{eq:cent_loss}
    \mathcal{L}_{CNT} =\frac{1}{N}\sum_{j=1}^N \left \| \mathbf{y}_j - \mu_{\pi(\mathbf{y}_j)} \right \|_2^2
\end{equation}

\noindent $\mu$ represents a centroid of a cluster $\pi$ taken from the spectral clustering output for a given sample $j$ \cite{zhang2019self}. Additionally, representation matrix \textbf{C} is forced to be symmetrical as it gives stable and unique solutions \cite{lu2018subspace}. Symmetric loss is defined as:

\begin{equation}
    \label{eq:symm}
    \mathcal{L}_{SYM} =\frac{1}{2}\sum^N_{j=1} \left \| \mathbf{c}_j - \mathbf{a}_j \right \|^2_2 = \frac{1}{2}\left \| \mathbf{C} -\mathbf{A} \right \|^2
\end{equation}

\noindent where \textbf{A} represents symmetric affinity matrix $\mathbf{A}=\frac{1}{2}(|\mathbf{C}|+|\mathbf{C}^T|)$. Every loss function has an assigned trade-off regularization constant $\lambda_1$ to $\lambda_6$ to regulate its importance. Thus, total loss function is defined as follows:

\begin{equation}
    \label{eq:total_loss}
    \mathcal{L}_{T} = \lambda_1 \mathcal{L}_{REC} + \lambda_2 \mathcal{L}_{CIM} + \lambda_3 \mathcal{L}_{CQ} + \lambda_4 \mathcal{L}_{CE} + \lambda_5 \mathcal{L}_{CNT} + \lambda_6 \mathcal{L}_{SYM}
\end{equation}

\noindent Also, to address the overfitting problem, we present results both on the train sets and on the independent test sets. The latter will give better insights into the generalization of used algorithms.

\section{Experimental results}
\label{sec:experimental}

In this section, we compare the clustering performance of Robust $S^2$ConvSCN with state-of-the-art self-supervised $S^2$ConvSCN model on four well-known datasets. Performance of Robust $S^2$ConvSCN model cannot be compared to the performance of shallow and recent deep subspace clustering methods reported in the literature, because such comparison would be unfair to the Robust $S^2$ConvSCN model. Shallow models use labels to optimize hyperparameters while deep models experience label-leakage as pretrained parts of their network use labels for early stopping (see Figure \ref{fig:acc_vs_loss}). In this research, the performance is evaluated in term of accuracy (Acc):

\begin{equation}
    Acc(\hat{\mathbf{r}},\mathbf{r})=\max_{\pi \in \Pi_k} \left (\frac{1}{N}\sum^N_{i=1}\left \{ \pi(\hat{\mathbf{r}_i}) = \mathbf{r}_i \right \} \right )
\end{equation}
\bigbreak
\noindent where $\Pi_k$ stands for the permutation space of [$k$]. We compare the performance of Robust $S^2$ConvSCN with state-of-the-art self-supervised deep subspace clustering algorithm\cite{zhang2019self}.

\subsection{Experimental setup}

Robust $S^2$ConvSCN model is implemented in Keras \cite{chollet2015keras} and Tensorflow \cite{tensorflow2015-whitepaper}. First, the random seed is fixed for reproducibility. To have a reliable performance estimate on the test set and multiple independent train and test folds for each dataset, a stratified k-fold splitting was performed. Depending on the dataset constraints, the train-test split was performed on every fold or one fold served as a train set while the rest $k-1$ folds served as a test set. Hence, independent $k$ observations of algorithms' performances for each dataset have been produced. 

As discussed in Section \ref{sec:robusts2convscn}, instead of monitoring accuracy every epoch, and consequently supervising the learning using ground truth labels, we use only the loss for reducing learning rate and early stopping. The training is stopped either after reaching the early stopping criterion or after reaching the maximum number of epochs. As in \cite{zhang2019self}, pretraining was also applied. Firstly, the autoencoder is pretrained to replicate the input. Secondly, the self-expression layer together with the pretrained autoencoder (DSCNet \cite{ji2017deep}) was trained to reach a limited number of epochs or the early stopping criterion. After that, the FC layer and self-supervision modules were added to the pretrained DSCNet. This procedure forms a starting point for both $S^2$ConvSCN and Robust $S^2$ConvSCN.

\begin{figure}[ht!]
    \centering
    \includegraphics[width=0.8\textwidth]{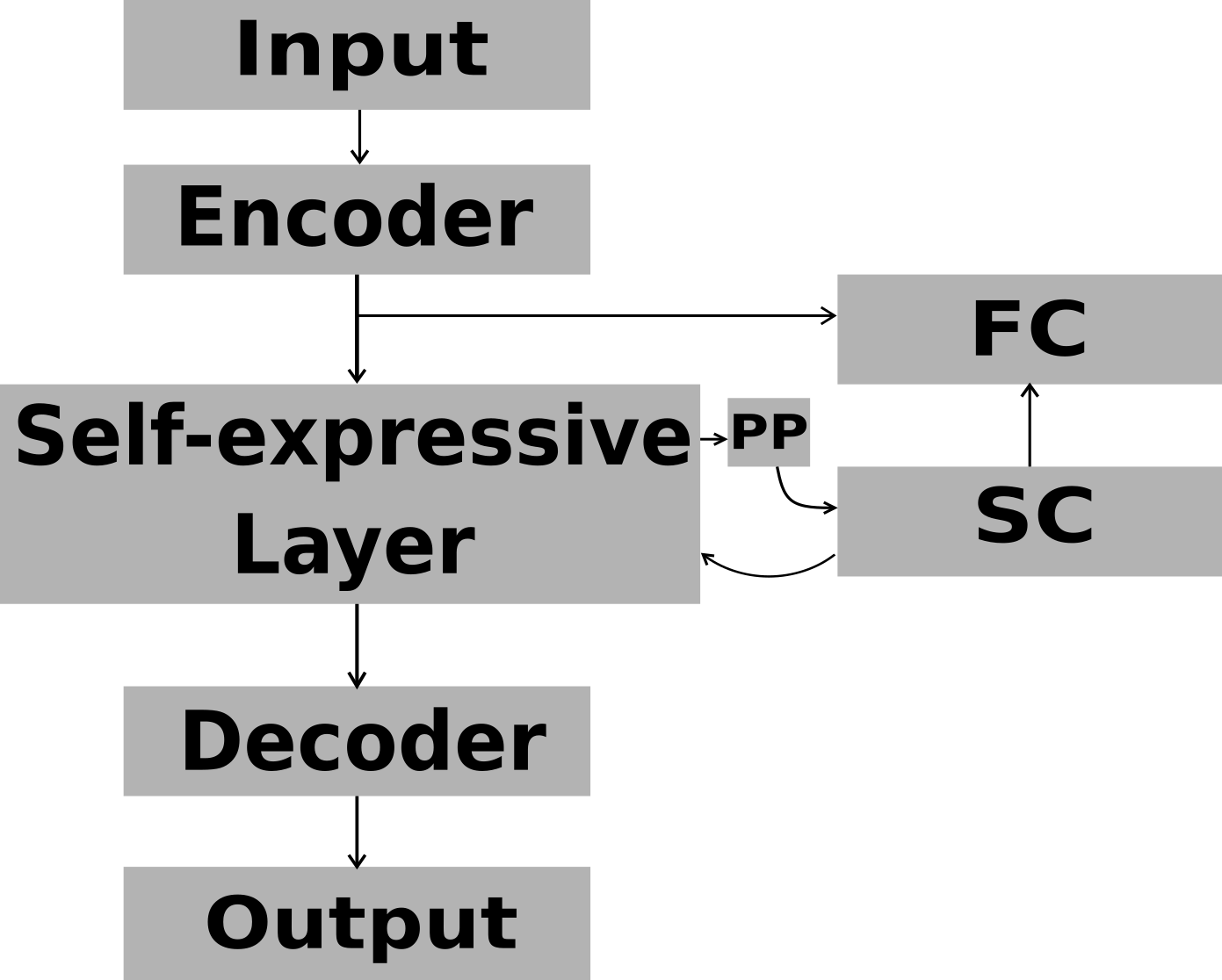}
    \caption{Flowchart representing the forward pass of Robust $S^2$ConvSCN training. Spectral clustering (SC) module serves as supervisor for FC and self-expressive layer. PP represents post-processing of representation matrix \textbf{C}.}
    \label{fig:flowchart}
\end{figure}

It is important to note that, once trained, the \textbf{C} matrix from the self-expression layer is post-processed in two steps, as done in DSCNet \cite{jiGithub}. First, it is thresholded by keeping only $\delta$ largest, in terms of magnitude, elements in a row. The accepted number of largest elements depends on their sum - $S_\delta$. If the sum $S_\delta$ exceeds $\lambda \sum^N_{i=0} c_{ji},\ i\in\{0,\dots,N \}$, where $\lambda$ stands for an empirically set thresholding constant and $N$ represents the number of elements in a row, other elements (not included in the sum $S_\delta$) are set to zero. Second, knowing the dimensionality of the dataset $d$, after SVD decomposition of thresholded representation matrix \textbf{C} only $d$ largest eigenvalues with related eigenvectors were kept meaning the remaining eigenvectors span the noise subspace. The reconstructed representation matrix serves as an input to the spectral clustering algorithm which produces pseudo-labels. Figure \ref{fig:flowchart} represents a flowchart of Robust $S^2$ConvSCN training.

During the warming-up phase, pseudo-labels were not refined because the newly introduced layers and modules could affect the convergence process (see Section \ref{sec:robusts2convscn} for the discussion). For the sake of fair comparison, different versions of $S^2$ConvSCN were trained using the same settings as Robust $S^2$ConvSCN. Table \ref{tab:dataset_setting} shows settings for MINST \cite{lecun2010mnist}, COIL-20 \cite{nene1996coil20}, COIL-100 \cite{nene1996coil100}, and Extended Yale B \cite{lee2005acquiring}. 

\begin{table}[ht]
\centering
\caption{Settings of Robust $S^2$ConvSCN per each dataset. T-max represents maximum number epochs and T0 stands for the number of epochs after which pseudo-labels are refined. Warm-up represents the number of epochs without the refinement of pseudo-labels, LR stands for starting learning rate, and Min LR represents the minimum learning rate.}
\label{tab:dataset_setting}
\begin{tabular}{l|r|r|r|r|}
\cline{2-5}
 & \multicolumn{1}{l|}{\textbf{MNIST}} & \multicolumn{1}{l|}{\textbf{COIL20}} & \multicolumn{1}{l|}{\textbf{COIL100}} & \multicolumn{1}{l|}{\textbf{EYaleB}} \\ \hline
\multicolumn{1}{|l|}{\textbf{data splits}} & 15 & 5 & 5 & 5 \\ \hline
\multicolumn{1}{|l|}{\textbf{T-max}} & 9000 & 4000 & 4000 & 9000 \\ \hline
\multicolumn{1}{|l|}{\textbf{T0}} & 30 & 50 & 40 & 30 \\ \hline
\multicolumn{1}{|l|}{\textbf{Warm-up}} & 50 & 100 & 80 & 50 \\ \hline
\multicolumn{1}{|l|}{\textbf{LR}} & 1e-3 & 1e-4 & 1e-4 & 1e-4 \\ \hline
\multicolumn{1}{|l|}{\textbf{Min LR}} & 1e-6 & 1e-6 & 1e-7 & 1e-6 \\ \hline
\end{tabular}
\end{table}

\subsection{Comparison on full datasets}
As previously discussed, training and testing on the same data is an overly optimistic estimation of the model's performance. Due to the fair comparison, we implemented and compared our model to the different versions of the $S^2$ConvSCN model in the same manner as reported in \cite{zhang2019self}. However, the lowest clustering error was not used as a criterion as it would be a violation of the unsupervised learning principle. Instead, a fixed number of epochs, learning rate decay, and early stopping based on loss decrease only were used. Thus, obtained results differ significantly from \cite{zhang2019self} report. For $S^2$ConvSCN, optimal regularization constants were transferred from \cite{zhang2019self}. In Table \ref{tab:comparison-full}, a comparison of $S^2$ConvSCN and Robust $S^2$ConvSCN with different \textbf{C} matrix regularization strategies can be seen. Due to memory limitations, MNIST could not be included in the comparison.

\begin{table}[ht]
\centering
\caption{Accuracy comparison of $S^2$ConvSCN and Robust $S^2$ConvSCN trained and tested on full COIL20, COIL100, and Extended YaleB datasets. BD and L2 stand for block diagonal and $\ell_2$ regularization of representation matrix, respectively.}
\label{tab:comparison-full}
\begin{tabular}{|l|r|r|r|}
\hline
 & \multicolumn{1}{c|}{\textbf{COIL20}} & \multicolumn{1}{c|}{\textbf{COIL100}} & \multicolumn{1}{c|}{\textbf{EYaleB}} \\ \hline
$\mathbf{RS^2}$\textbf{ConvSCN + BD} & 0.76250 & 0.50528 & 0.41159 \\ \hline
$\mathbf{RS^2}$\textbf{ConvSCN + L2} & \textbf{0.88403} & \textbf{0.68805} & \textbf{0.81661} \\ \hline
$\mathbf{S^2}$\textbf{ConvSCN + BD} & 0.81111 & 0.30375 & 0.56867 \\ \hline
$\mathbf{S^2}$\textbf{ConvSCN + L2} \cite{zhang2019self} & 0.63333 & 0.55319 & 0.75247 \\ \hline
\end{tabular}
\end{table}

\subsection{Classification performance on unseen data}

Having several independent observations is crucial for the model's performance validation. Thus, according to Table \ref{tab:dataset_setting}, datasets were split to stratified folds. For MNIST, as there is enough data, each fold was additionally divided into a train and test set, 70\% and 30\%, respectively. For other datasets, the remaining folds were all used for testing as these datasets have a very limited number of data per label, i.e. additional splitting as for MNIST could not be possible. Testing results for all datasets can be seen in Table \ref{tab:comparison-independent}.

\begin{table}[ht]
\centering
\caption{Accuracy comparison of $S^2$ConvSCN and Robust $S^2$ConvSCN trained and tested on independent folds for each dataset. BD and L2 stand for block diagonal and $\ell_2$ regularization of representation matrix, respectively. The best mean accuracy for each dataset is bolded.}
\label{tab:comparison-independent}
\resizebox{0.8\textwidth}{!}{%
\begin{tabular}{|l|l|r|r|r|r|}
\hline
 &  & \multicolumn{1}{l|}{\textbf{MNIST}} & \multicolumn{1}{c|}{\textbf{COIL20}} & \multicolumn{1}{c|}{\textbf{COIL100}} & \multicolumn{1}{c|}{\textbf{EYaleB}} \\ \hline
\multirow{2}{*}{$\mathbf{RS^2}$\textbf{ConvSCN + BD}} & \textbf{mean} & \textbf{0.50417} & 0.79824 & 0.44882 & 0.50017 \\ \cline{2-6} 
 & \textbf{stddev} & 0.05329 & 0.02984 & 0.03845 & 0.02695 \\ \hline
\multirow{2}{*}{$\mathbf{RS^2}$\textbf{ConvSCN + L2}} & \textbf{mean} & 0.42979 & \textbf{0.82547} & \textbf{0.55461} & \textbf{0.75239} \\ \cline{2-6} 
 & \textbf{stddev} & 0.04878 & 0.02537 & 0.01454 & 0.03346 \\ \hline
\multirow{2}{*}{$\mathbf{S^2}$\textbf{ConvSCN + BD}} & \textbf{mean} & 0.12659 & 0.71686 & 0.33762 & 0.49187 \\ \cline{2-6} 
 & \textbf{stddev} & 0.06258 & 0.07814 & 0.04736 & 0.04261 \\ \hline
\multirow{2}{*}{$\mathbf{S^2}$\textbf{ConvSCN + L2} \cite{zhang2019self}} & \textbf{mean} & 0.12659 & 0.67486 & 0.52099 & 0.63735 \\ \cline{2-6} 
 & \textbf{stddev} & 0.02569 & 0.05917 & 0.02295 & 0.07062 \\ \hline
\end{tabular}
}
\end{table}

\subsection{Complexity and convergence}

Dataset size and choice of representation matrix regularization have a direct impact on time and memory complexity. As the model must learn in batch mode, the number of samples in the dataset $N$ determines the number of parameters in the self-expression layer - $N^2$. Thus, dataset size impacts memory and time complexity. Also, regularization imposed on the representation matrix affects the computational time of the algorithm. As presented in Section \ref{sec:prelim}, BD regularization is defined as the sum of $k$ smallest eigenvalues of Laplacian matrix \textbf{L}. Thus, if the BD regularization is chosen, the model will have higher time complexity due to the usage of the singular value decomposition algorithm. Although it is not directly associated with the time complexity per epoch, the spectral clustering module refines pseudo-labels every $T0$ epochs which adds to overall training time.

To properly emphasize individual loss functions, all regularization constants in the overall loss function must be tuned (discussed in Section \ref{sec:robusts2convscn}). As a good starting point, we used regularization constants from \cite{zhang2019self} and refined only those associated with the error measure term in the self-expression layer and representation matrix \textbf{C}. Depending on the fold size and regularization used, training of one fold on one GPU takes 2-6 hours. Testing on one fold is within several seconds.

\section{Discussion and conclusion}
\label{sec:discussion}

As it could be seen in Table \ref{tab:comparison-independent}, Robust $S^2$ConvSCN outperforms $S^2$ConvSCN on all datasets regardless of regularization imposed on representation matrix \textbf{C}. Moreover, in the case of the MNIST dataset, Robust $S^2$ConvSCN with BD regularization leads to the best results. The CIM loss, indeed, handles the corruptions in the data better than MSE, especially the possible amplification of corruptions due to their propagation through deep layers. 

However, training of deep learning algorithms is an especially time-consuming task. Thus, settings in Table \ref{tab:dataset_setting} and regularization constants we found could possibly be improved in further testing. As authors of the $S^2$ConvSCN have found optimal settings for their model, and having in mind that these settings are optimized violating the unsupervised learning principles previously discussed, we assume that their approach leads to an overly optimistic estimation of model's performance. Nevertheless, Robust $S^2$ConvSCN outperformed $S^2$ConvSCN showing its superiority. 

Using datasets for which the dimensionality $d$ is unknown could be a challenging task. Without that information, it is difficult to post-process the representation matrix to eliminate the noise. However, avoiding a need for any \textit{a priori} knowledge is an important task to deal with in further research. Also, there is a problem with pseudo-labels refinement frequency. If it occurs too often it could lead to divergence of the whole model. If the refinement occurs too rarely, the model could get stuck in the poor minimum. Finding a more robust way of pseudo-label refinement is still an open question. 

A possible problem in these algorithms could be a disconnection of overall loss decrease on one hand and accuracy increase on the other. As there are many loss functions in Robust $S^2$ConvSCN, not every loss is equally important for the correct final classification of given samples. Thus, observing overall loss for early stopping and learning rate reduction might be suboptimal. More research towards identifying which loss has higher significance during the training is needed.

As some often used datasets for comparison \cite{nene1996coil20,nene1996coil100,lee2005acquiring} have a small number of data per class, it is not good to have a lot of parameters in the network. However, for datasets like \cite{lecun2010mnist}, the problem is the opposite. As there are a lot of samples in the dataset, the model is reaching a memory constraint due to the size of the representation matrix \textbf{C} trained in the self-expression layer. In that sense, mini-batch training of deep subspace clustering models, such as one presented in this study, could cross the barrier of memory complexity and offer more approachable learning.

To conclude, among finding a more robust model, this study aimed to set up a new, more transparent way for the evaluation of deep subspace clustering models. Majority subspace clustering studies optimize hyperparameters based on the data the algorithm is trained and tested on. From the unsupervised learning perspective, it is unacceptable to tune hyperparameters based on true labels. Moreover, algorithms like \cite{zhang2019self} suffer from an early commitment problem as they depend on weight-transfer from pretrained models, e.g. \cite{jiGithub}, which have "seen" the true labels already. For that, we propose an evaluation of the algorithm on independent data to have a proper estimate of model performance. As can be seen from Table \ref{tab:comparison-independent}, measured performances of $S^2$ConvSCN significantly differ from the optimistic one presented in \cite{zhang2019self}. Additionally, the presented model which incorporates label-free learning and the robust correntropy loss can easily be extended to multi-modal and multi-view data. We aim to address that in our further research. 

\section*{Acknowledgments}
This work has been supported by the Croatian Science Foundation Grant IP-2016-06-5235 and in part by the European Regional Development Fund (DATACROSS) under Grant KK.01.1.1.01.0009. We gratefully acknowledge the support of NVIDIA Corporation with the donation of the QUADRO P6000 GPU used for this research.

\bibliography{main}

\end{document}